\documentclass{article}
\usepackage[T1]{fontenc}
\usepackage{iclr2027_conference,times}

\usepackage{amsmath,amsfonts,bm}

\def\eqref#1{equation~\ref{#1}}

\def\1{\bm{1}}

\DeclareMathAlphabet{\mathsfit}{\encodingdefault}{\sfdefault}{m}{sl}
\SetMathAlphabet{\mathsfit}{bold}{\encodingdefault}{\sfdefault}{bx}{n}

\DeclareMathOperator*{\argmin}{arg\,min}

\usepackage{amsmath,amssymb}
\usepackage{booktabs}
\usepackage{graphicx}
\usepackage{microtype}
\usepackage[table]{xcolor}
\usepackage{multirow}
\usepackage{tabularx}
\usepackage{float}
\usepackage{flafter}
\usepackage{placeins}
\usepackage{algorithm}
\usepackage{algpseudocode}
\usepackage{url}
\usepackage{hyperref}
\usepackage{xspace}

\hypersetup{
  hidelinks,
  pdftitle={When Should Agents Check External State? Budgeting Observations for Stored Intentions},
  pdfauthor={Zhengkun Di, Bin Shi, Kai Sun, Yiming Xu, and Bo Dong}
}

\definecolor{oursblue}{HTML}{0072B2}
\definecolor{staticrow}{HTML}{E7F4F2}
\newcommand{\method}{\textsc{BudgetPM}\xspace}
\newcommand{\due}{Matched Due\xspace}
\newcommand{\best}[1]{\textbf{#1}}

\title{When Should Agents Check External State?\\
Budgeting Observations for Stored Intentions}

\author{Zhengkun Di$^{1}$, Bin Shi$^{1,*}$, Kai Sun$^{1}$,
Yiming Xu$^{1}$, Bo Dong$^{2}$\\
\normalfont$^{1}$School of Computer Science and Technology, Xi'an Jiaotong University, Xi'an, China\\
$^{2}$School of Distance Education, Xi'an Jiaotong University, Xi'an, China\\
\texttt{dizhengkun@stu.xjtu.edu.cn},
\texttt{shibin@xjtu.edu.cn}\\
\texttt{xym0924@stu.xjtu.edu.cn},
\texttt{dong.bo@xjtu.edu.cn}\\
$^{*}$Corresponding author: \texttt{shibin@xjtu.edu.cn}}

\iclrfinalcopy

\begin{document}
\maketitle
\fancyhead[L]{Preprint}

\begin{abstract}
Prospective memory allows an agent to retain an intention tied to a future
condition, but the stored intention does not reveal whether that condition
currently holds. Checking it may require web access, multi-step tool use, and
paid calls. Existing systems decide when intentions require attention, but do
not allocate the resulting observations under a shared budget. We introduce
the first resource-allocation formulation for the external observations
required by stored intentions under a shared episode budget. BudgetPM offers two
policy variants that share a hard-budget executor. BudgetPM-Static uses a
lightweight Logistic scorer to learn whether a check improves the current
decision. BudgetPM-Sequential distills
full-episode hindsight schedules into a lightweight policy that decides when
to spend or reserve capacity using only pre-query information at deployment.
We evaluate BudgetPM against two
public memory-agent systems, five matched controls, and four hand-designed
monitoring or budget-adaptation rules. Across two benchmarks and three
backbones, BudgetPM-Static outperforms adapted Mem0 and PMA workflows. On
PM-Bench, its Logistic scorer reaches competitive quality--cost operating
points alongside higher-capacity scorers and retains 99.9--100\% of
unconstrained quality with 42--54\% fewer observations. Under severe scarcity
and the same hard caps, BudgetPM-Sequential exceeds the strongest tested
natural monitoring schedule by 1.92--2.58 Set F1 points. It reaches the same
Set F1 and on-time recall with 16--33\% fewer observations. Matched
attribution, exact-cost analysis, and a fixed-budget load
intervention link this gain to competition between present and future
opportunities. These results yield a demand--capacity design rule: local gating works when
capacity covers demand, while future-aware supervision adds value when
observations compete across time.
\end{abstract}

\section{Introduction}
Long-running agents often retain instructions whose execution depends on
external state. Consider ``Release the order when the inspection status changes
to passed in the supplier portal.'' The agent can remember this instruction,
but it must still log in to the portal and inspect its current state to
determine whether the condition has been met. A status check may involve web
access, authentication, several tool calls, model inference, or paid API
access. Repeated checks can therefore incur substantial interaction and
computational costs~\citep{kapoor2025mdps,liu2025budget}.

As an agent retains more intentions, they create recurring demands for fresh
information. Recent work brings prospective memory into agent systems, where
agents retain intentions for future action and act when the corresponding
conditions are satisfied~\citep{liu2026pm,wu2026remember,zhang2026triggerbench}.
These systems use predefined rules or prompted agents to decide when an
intention requires attention and when more evidence should be requested
\citep{zhao2026making}. They do not, however, allocate repeated external checks
when those checks compete for limited observation capacity.

Fixed-interval polling, more frequent checks near an expected trigger, and
backoff after unchanged observations are natural solutions. They may work well
when trigger times are predictable or capacity is ample. When event timing
varies, fixed intervals may delay detection. Checks concentrated near an
expected trigger may miss early changes or waste calls when the event is
delayed. Backoff may postpone detection after a long unchanged period. More
fundamentally, each rule relies on a preset cadence, an expected trigger time,
or past observations. None directly compares the value of checking now with
the value of preserving capacity for future opportunities. This leads to two
questions: \emph{which observations are useful
now, and when should an agent reserve capacity for future checks?}

We introduce \method with two policy variants and a shared
hard-budget executor. Static learns whether an observation improves the current
decision. Sequential learns from full-episode hindsight schedules how to spend
or reserve capacity across time. Full-episode information is used only for
offline supervision. At deployment, both policies use the observable pre-query
state, and the executor enforces the per-step and episode budgets.

Our experiments evaluate BudgetPM at three levels. Complete-system comparisons
span two benchmarks and three backbones, where Static achieves the highest
end-to-end task quality. Matched local-selection tests show that its lightweight
Logistic scorer reaches competitive quality--cost operating points alongside
LambdaMART and a fixed ensemble. Static also preserves 99.9--100\% of
unconstrained PM-Bench quality with 42--54\% fewer observations. Temporal tests
compare Sequential with four hand-designed rules and one matched future-blind
control. Under the same hard caps, Sequential exceeds the strongest natural
schedule by 1.92--2.58 Set F1 points. It uses 16--33\% fewer observations to
match the schedule's Set F1 and on-time recall. Budget
sweeps and a fixed-budget load intervention link these gains to competition
between present and future checks.

Our contributions are:
\begin{itemize}
\setlength\itemsep{1pt}
\item \textbf{Budgeted external-state checks.} We formulate how the observations
required by stored intentions should share an episode budget.
\item \textbf{Lightweight future-aware allocation.} \method combines local-value
estimation, offline schedule supervision, and hard-budget execution. Sequential
distills full-episode schedules into a lightweight deployment policy.
\item \textbf{A demand--capacity boundary.} Natural schedules, matched temporal
policies, exact-cost analysis, and a fixed-budget intervention show when
future-aware allocation adds value beyond local selection.
\end{itemize}

\section{Related Work}
\paragraph{Memory systems and prospective triggers.}
Memory-augmented agents preserve experience through reflection, virtual
context, structured stores, and learned memory operations
\citep{park2023generative,shinn2023reflexion,packer2023memgpt,chhikara2025mem0,xu2026mem,yu2026agentic}.
Recent systems improve the efficiency of memory construction, retrieval,
summarization, and module selection
\citep{zhang2026lightweight,liao2026leanmem,zhang2026learning,lei2026memorycpt}.
Prospective-memory work instead asks when a stored intention becomes relevant.
PMA decides when to surface a reminder~\citep{wu2026remember}, and TriggerBench tests
whether an agent responds to triggering evidence in its context
\citep{zhang2026triggerbench}. PM-Bench includes triggers that require active
queries to hidden state~\citep{liu2026pm}; PIS uses structured rules and a
small language model to decide whether such a query is needed
\citep{zhao2026making}. LongMemEval and MemoryArena broaden long-term memory
evaluation~\citep{wu2024longmemeval,he2026memoryarena}, while OSL-MR allocates
storage or context across retained memories~\citep{kang2026learning}. These
systems decide when an intention needs attention and what evidence may resolve
it. BudgetPM addresses the next allocation problem: how repeated external-state
checks should share limited capacity.

\paragraph{Budget-aware external interaction.}
FrugalGPT, RouteLLM, and WISERouter allocate computation under cost or workload
constraints~\citep{chen2023frugalgpt,ong2025routellm,li2026wiserouter}. BATS
uses explicit tool-call budgets. CostBench evaluates cost-aware multi-turn tool
planning, and When2Tool learns when calls are useful
\citep{liu2025budget,liu2026costbench,sun2026llm}. These works show that
external interaction can be limited by cost, workload, or call quota. BudgetPM
connects these lines of work by treating the external-state checks required by
stored intentions as consumers of a shared call budget.

\paragraph{Sequential observation allocation.}
State-sensing and value-of-information work asks when state is worth its
acquisition cost~\citep{kapoor2025mdps,krause2009optimal}. Adaptive submodularity
and active feature acquisition formalize sequential information gathering
\citep{golovin2011adaptive,shim2018joint}. Bandits with knapsacks and constrained
MDPs link decisions through consumable resources
\citep{badanidiyuru2018bandits,jain2022towards}. INTENT plans future priced tool use
with a learned world model~\citep{liu2026budget}. These formulations explain why
spending capacity now can reduce the options available later. BudgetPM asks
when current-value selection is sufficient and when limited capacity requires
allocation across time. It constructs full-episode schedules offline
\citep{rusu2015policy} and distills them into a lightweight deployment policy.

\section{BudgetPM: Allocating External-State Checks}
\begin{figure}[h]
  \centering
  \includegraphics[width=\linewidth]{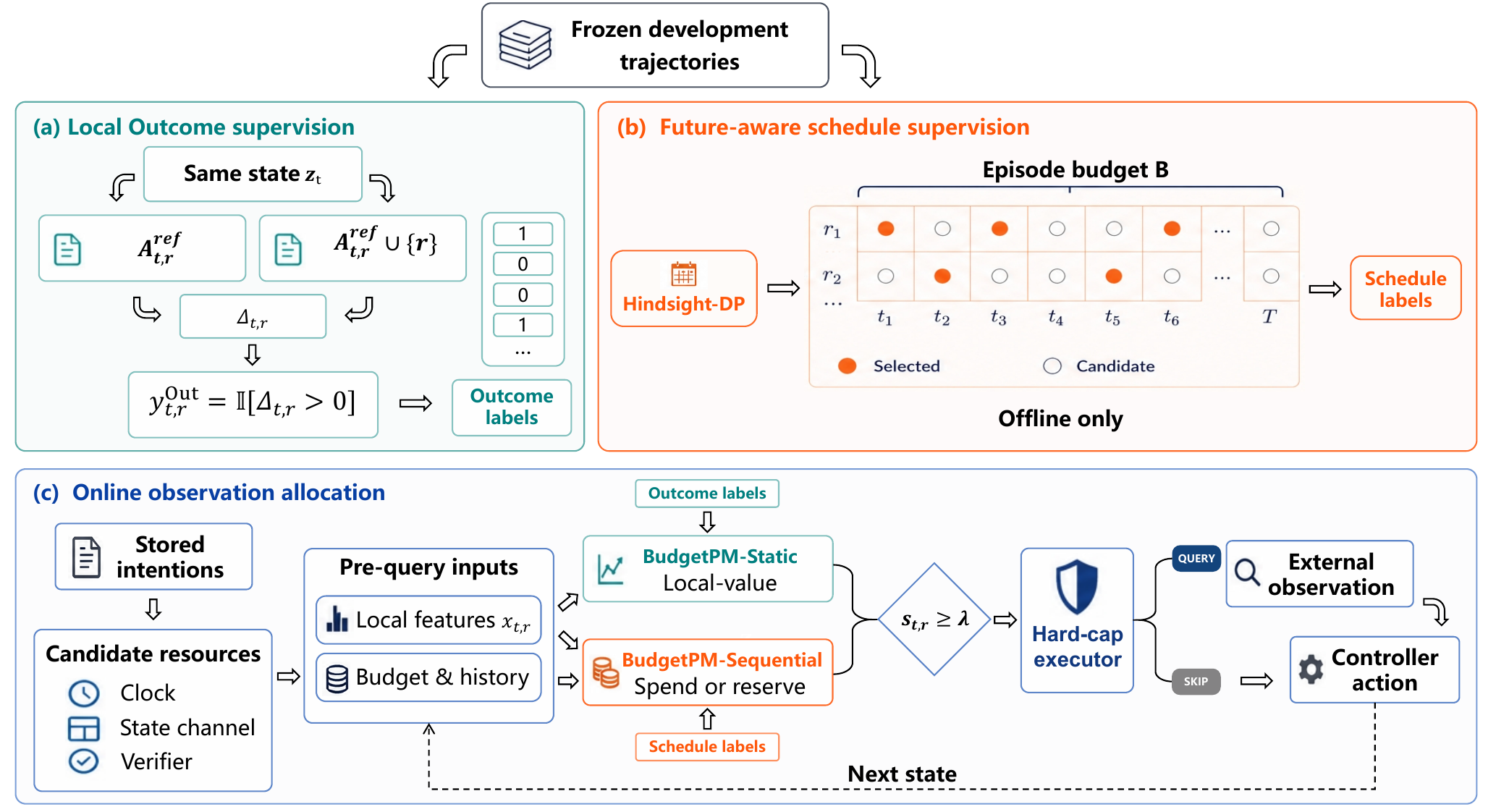}
  \caption{Overview of BudgetPM. Frozen development trajectories provide cached
  observation results for constructing (a) local Outcome labels and
  (b) hindsight-DP schedule labels under an episode budget. At deployment
  (c), Static or Sequential uses only pre-query inputs; thresholded proposals
  pass through a shared hard-cap executor that enforces the per-step and episode
  limits. Future information is used only for offline supervision.}
  \label{fig:framework}
\end{figure}

BudgetPM receives active intentions from an upstream memory system and decides
which external-state resources to query under the available budget.
Figure~\ref{fig:framework} shows how frozen trajectories provide local Outcome
labels for Static (panel~a) and hindsight-DP schedule labels for Sequential
(panel~b). At deployment (panel~c), either policy uses only pre-query inputs,
and a shared executor enforces the per-step and episode budgets.

\subsection{Problem formulation}
\label{sec:method_online}

Consider an episode $\xi$ of $T$ steps. Before step $t$, the history $h_t$
contains the active intentions and all observations collected so far. These
intentions identify queryable external-state resources $\mathcal R_t$, such as
a clock, calendar, package-status provider, or semantic verifier. Because one
resource may serve several intentions, a single query can resolve several
triggers. Querying $r\in\mathcal R_t$ costs a known amount $c(r)>0$.

An online policy $\pi$ uses $h_t$ to choose a query set
$S_t^{\pi}(\xi)\subseteq\mathcal R_t$. Across the episode, these choices
produce a trajectory $\tau_{\xi}(\pi)$, whose task quality is measured by
$M(\tau_{\xi}(\pi))$. Under the episode-level hard budget $B$, the objective is
\begin{equation}
\begin{aligned}
\max_{\pi}\quad &\mathbb E_{\xi}[M(\tau_{\xi}(\pi))]\\
\text{s.t.}\quad
&\sum_{t=1}^{T}\sum_{r\in S_t^{\pi}(\xi)}c(r)\le B
\quad\forall\xi,
\end{aligned}
\label{eq:objective}
\end{equation}
where the expectation is over episodes and provider or controller randomness.
BudgetPM fits two variants from frozen development trajectories: Static uses
labels of current-decision improvement, whereas Sequential uses labels from
budget-constrained full-episode schedules. Both are lightweight scorers and
share the same hard-cap executor.

\subsection{BudgetPM-Static: Local Outcome supervision}
\label{sec:method_static}

Static asks whether an available external-state check would improve the current
decision. Retrieval may associate several intentions with the same resource,
so the policy scores resources rather than individual memories.

At each recorded step, a frozen trajectory provides the pre-query controller
state $z_t$ and cached outcomes for a candidate set
$A_t\subseteq\mathcal R_t$. For any $A\subseteq A_t$, we insert its cached
evidence into $z_t$ and recompose the action $a_t(A;z_t)$. The step-level metric
$M_t$ scores this action, whereas $M$ in Equation~\ref{eq:objective} scores the
episode. For compactness, define
\[
m_t(A):=M_t(a_t(A;z_t)).
\]
For each candidate resource $r\in A_t$, let
$A^{\mathrm{ref}}_{t,r}\subseteq A_t\setminus\{r\}$ be a comparison set that
excludes $r$. The frozen local Outcome effect is the score change caused by
adding $r$ to this same set:
\begin{equation}
\Delta_{t,r}
=
m_t\!\left(A^{\mathrm{ref}}_{t,r}\cup\{r\}\right)
-
m_t\!\left(A^{\mathrm{ref}}_{t,r}\right),
\qquad
y^{\mathrm{Out}}_{t,r}
=
\mathbb I[\Delta_{t,r}>0].
\label{eq:local_effect}
\end{equation}
Here $\mathbb I[\cdot]$ is the indicator function. Thus,
$y^{\mathrm{Out}}_{t,r}=1$ when evidence from $r$ improves the current decision
under the comparison set. Recomposition keeps $z_t$ fixed and only creates
training labels; at deployment, each policy updates its own budget, query
history, and task state.

\subsection{BudgetPM-Sequential: Future-aware schedule supervision}
\label{sec:method_temporal}

Local Outcome labels measure current value, but not when the episode budget
should be spent. Sequential learns this timing from schedules optimized over
the complete frozen episode. Its hindsight-DP teacher uses dynamic
programming~\citep{bellman1966dynamic} to maximize the frozen episode score
under budget $B$.

\paragraph{Episode-level schedule optimization.}
A schedule $S_{1:T}=(S_1,\ldots,S_T)$ selects $S_t\subseteq A_t$ at each step
within the per-step limits. Let $Q_\xi(S_{1:T})$ be the frozen episode score
produced by all selected evidence. The teacher solves
\begin{equation}
S^{\star}_{1:T}
\in
\arg\max_{S_t\subseteq A_t}\ Q_\xi(S_{1:T})
\quad
\text{s.t.}\quad
\sum_{t=1}^{T}\sum_{r\in S_t}c(r)\le B.
\label{eq:schedule_quality}
\end{equation}
Dynamic programming extends partial schedules one step at a time. At each
cumulative cost, it removes a state if another state is at least as good on
every statistic used by $Q_\xi$. A fixed rule breaks exact ties. The maximizing
schedule defines the binary target
$y^{\mathrm{Seq}}_{t,r}=\mathbb I[r\in S_t^{\star}]$ for each candidate.
Full-episode information is used only to construct these offline targets.

\subsection{Policy fitting and online allocation}
\label{sec:method_deployment}

Both variants share the same scoring-and-execution pipeline. Static uses local
pre-query features; Sequential adds nine observable budget and history
features, including remaining budget, episode progress, and recent query usage.
We denote either feature vector by $x_{t,r}$. After standardization, a
class-balanced logistic regression model $f_\theta$ produces the score
$s_{t,r}=f_\theta(x_{t,r})$~\citep{pedregosa2011scikit}. Static is trained on
$y^{\mathrm{Out}}_{t,r}$, whereas Sequential is trained on
$y^{\mathrm{Seq}}_{t,r}$. Model fitting uses development trajectories only.

At each step, the active intentions determine $\mathcal R_t$. Before any
external call, BudgetPM constructs $x_{t,r}$ from $h_t$ and scores every
candidate. Those with $s_{t,r}\ge\lambda$, where $\lambda$ is the operating
threshold, form the proposal set $P_t$. The hard-cap executor selects a feasible
subset under the remaining episode budget, per-step cost cap $L$, and per-step
query-count cap $N_{\max}$, as shown in
Algorithm~\ref{alg:online_allocation}. The selected set is fixed before any
provider is called, so neither scoring nor selection uses the current query
results.

\begin{algorithm}[h]
\caption{BudgetPM online observation allocation}
\label{alg:online_allocation}
\small
\begin{algorithmic}[1]
\Require Trained scorer $f_\theta$, operating threshold $\lambda$, episode cap $B$,
\Statex \hspace{\algorithmicindent} per-step cost cap $L$, per-step query-count cap $N_{\max}$, horizon $T$
\State Initialize remaining budget $B_{\mathrm{rem}}\gets B$ and observable history $h_1$
\For{$t=1,\ldots,T$}
  \State Generate distinct candidate resources $\mathcal R_t$ from active intentions in $h_t$
  \State Build all pre-query features; score $s_{t,r}\gets f_\theta(x_{t,r})$
  \State Form proposals $P_t\gets\{r\in\mathcal R_t:s_{t,r}\ge\lambda\}$
  \State Sort $P_t$ by decreasing $(s_{t,r}/c(r),s_{t,r})$, then increasing resource ID
  \State Initialize selected set $S_t\gets\varnothing$ and step cost $u\gets0$
  \For{each $r$ in the sorted proposals}
    \If{$u+c(r)\le\min(B_{\mathrm{rem}},L)$ and $|S_t|<N_{\max}$}
      \State $S_t\gets S_t\cup\{r\}$;\quad $u\gets u+c(r)$
    \EndIf
  \EndFor
  \State Commit the plan and charge its cost: $B_{\mathrm{rem}}\gets B_{\mathrm{rem}}-u$
  \State Query only $S_t$ and pass the returned evidence to the controller
  \State Execute the controller action and update $h_{t+1}$, including spend history
\EndFor
\end{algorithmic}
\end{algorithm}

The executor skips an unaffordable proposal and continues scanning the list.
The query-count cap is $N_{\max}=\infty$ when unset.

\paragraph{Threshold calibration.}
We select $\lambda$ from out-of-fold predictions on the development set $D$.
Under unit costs, let $C_D(\lambda)$ count queries after per-step filtering and
before the episode cap; $C_D^{\max}$ is the count when all candidates are
proposed. For target budget fraction $b$, define
\[
K_b=\operatorname{round}\!\left(bC_D^{\max}\right).
\]
Let $\Lambda_D^{\mathrm{OOF}}$ be the set of distinct out-of-fold scores on
$D$. We choose
\begin{equation}
\lambda_b
=
\argmin_{\lambda\in\Lambda_D^{\mathrm{OOF}}}
\left|C_D(\lambda)-K_b\right|,
\label{eq:calibration}
\end{equation}
with ties favoring the higher threshold. The threshold controls the proposal
rate; the shared executor guarantees the per-step and episode limits.
Feature and calibration details are in Appendix~\ref{app:implementation}. The
matched label control in Section~\ref{sec:supervision_control} tests the
contribution of Outcome supervision.

\subsection{Exact-cost structural headroom}
Structural headroom asks whether future-aware scheduling can improve the
full-episode outcome at the same cost. Current-greedy selects the feasible
subset with the highest present-step frozen-recomposition score; let
$S^{\mathrm{CG}}_{1:T}$ denote its schedule. On the same frozen episode,
hindsight DP produces $S^{\mathrm{DP}}_{1:T}$ at the exact integer query cost
used by current-greedy. Define
\[
Q_{\mathrm{CG}}
:=
Q_\xi(S^{\mathrm{CG}}_{1:T}),
\qquad
Q_{\mathrm{DP}}
:=
Q_\xi(S^{\mathrm{DP}}_{1:T}).
\]
Their quality difference is the \emph{structural headroom}
\[
H
=
Q_{\mathrm{DP}}-Q_{\mathrm{CG}}.
\]
A positive $H$ means that a better full-episode allocation exists at the same
cost. Both teachers use the same current-step outcomes; only DP compares future
opportunities. An additional analysis fixes per-step cost to separate
allocation across time from resource choice within a step
(Appendix~\ref{app:headroom_decomposition}). Thus, $H$ diagnoses temporal
allocation headroom; learned-policy gains are evaluated separately.

\section{Experimental Setup}
\paragraph{Evaluation overview.}
We evaluate BudgetPM at three levels: complete-system comparisons measure
end-to-end effectiveness; matched controls isolate local and temporal
supervision; exact-cost teachers and a fixed-budget load intervention test when
allocation across time can improve quality.

\paragraph{Benchmarks and metrics.}
PM-Bench~\citep{liu2026pm} provides seven-day streams with visible, temporal,
and latent-state prospective-memory cues. We budget clock and named
state-channel queries and report episode Set F1, which also serves as
$Q_\xi$ in Equation~\ref{eq:schedule_quality}. Temporal experiments report
on-time recall, the fraction of reference triggers resolved before they become
late.
GoodAI LTM~\citep{castillo2024beyond} provides prospective-memory and
trigger-response tasks. We gate its semantic-verifier calls and report macro
accuracy over the two components.

\paragraph{Observation costs and budgets.}
We count state-provider or semantic-verifier calls as interaction cost. B$q$
denotes an episode cap set to approximately $q\%$ of the mean number of
feasible queries on the development set, rounded to an integer. For PM-Bench,
B5/B10/B15/B20/B40/B60/B80 permit
10/21/31/42/84/126/168 queries, based on an all-relevant mean cost of 209.4.
GoodAI B20/B40/B60/B80 permit 10/21/31/41 queries, based on an all-semantic
mean cost of 51.5. PM-Bench temporal experiments use unit costs and a per-step
cap of two queries.

\paragraph{Supervision construction.}
PM-Bench fixes scenario timing and provider events independently of query
decisions, so all policies face the same opportunity sequence. Its Outcome
labels use the singleton-versus-empty effect
($A^{\mathrm{ref}}_{t,r}=\varnothing$); GoodAI uses the leave-one-out effect
($A^{\mathrm{ref}}_{t,r}=A_t\setminus\{r\}$). For schedule supervision, the
PM-Bench DP retains cumulative $(\mathrm{TP},\mathrm{FP})$ states at each
integer cost and removes states dominated in both statistics.

\paragraph{Budget regimes.}
We use B20--B80 to study local selection on both benchmarks and PM-Bench
B5--B20 to locate the onset of temporal competition. Static is calibrated at
B20--B80 and reuses its frozen B20 threshold at B5/B10. Myopic and Sequential
are calibrated separately for their target caps. Temporal experiments use unit
costs to isolate allocation. Heterogeneous-price results appear in
Appendix~\ref{app:reviewer_diagnostics}.

\paragraph{Complete-system comparison.}
The complete-system comparison measures end-to-end task quality under each
system's native workflow. At B60, BudgetPM-Static is compared with
Mem0~\citep{chhikara2025mem0} and PMA~\citep{wu2026remember} across 30 seeds and
three backbones: Llama-3-8B, Qwen2.5-14B, and
Qwen3-14B~\citep{grattafiori2024llama,qwen2025qwen25,yang2025qwen3}. Subsequent
matched tests share the candidate space, controller, and executor and use
Llama-3-8B unless stated otherwise. Full details appear in
Appendix~\ref{app:system_baselines}.

\paragraph{Local-gating controls.}
\label{sec:supervision_control}
All controls use the same candidate-query space, episode cap, and hard-cap
executor. They vary either the selection signal or the scorer.
\emph{Relevance} directly uses PM-Bench retrieval relevance; GoodAI uses the
corresponding trigger-text similarity. Random-feasible and unconstrained
references appear in Appendix~\ref{app:random_references}. To test the value of
the supervision target, we define the \due label
$y^{\mathrm{Due}}_{t,r}=\mathbb I[r\text{ is needed to resolve a trigger due at }t]$.
It marks current trigger need, not the observation's effect on task score. The
Outcome-trained Static gate and \due share the same features,
Logistic model, development data, calibration, ordering, and executor. Only the
supervision label changes. To test whether Outcome supervision requires a more
expressive scorer, LambdaMART~\citep{burges2010ranknet} and a fixed
hurdle/pairwise \emph{Ensemble} replace the Logistic scorer while retaining the
same Outcome-supervised workflow.

\paragraph{Future-aware supervision controls.}
Myopic isolates future-aware supervision. It imitates an offline current-greedy
teacher that evaluates current-step outcomes, whereas Sequential imitates
hindsight DP, which can also compare future opportunities. Both policies use
the same 36 pre-query features, Logistic model, fitting procedure, grouped
out-of-fold protocol, threshold grid, ordering rule, and hard-cap executor.
They are evaluated on paired runs of the same task traces. Sequential
minus Myopic therefore measures the contribution of future-aware supervision.
Exact-cost hindsight DP minus current-greedy measures structural headroom.

\paragraph{Natural monitoring baselines.}
\emph{Periodic-Outcome}, \emph{Deadline-Outcome}, and
\emph{Deadline+Backoff-Outcome} share Static's frozen Outcome scorer and differ
only in how they distribute queries over time.
\emph{Periodic-Outcome} uses a uniform cadence. \emph{Deadline-Outcome}
prioritizes clock checks when an unresolved intention enters its visible time
window, and \emph{Deadline+Backoff-Outcome} additionally reduces state-channel
polling after unchanged observations. \emph{Pacing-LB} adjusts Static's
threshold using the remaining budget and horizon; its B5/B10 thresholds are
selected on development weeks. The natural schedules test deployable timing
rules, while Myopic isolates the value of future-aware supervision.

Within each temporal test, all policies are evaluated on the same task trace
and share the per-step limit, episode cap, and hard-cap executor. The three
natural schedules are swept over 20 caps to form Set F1 and on-time recall
curves against realized observation cost. Pacing-LB is evaluated at B5/B10. These
comparisons ask whether explicit timing rules can match Sequential's Set F1 and
on-time recall with comparable observation use.

\paragraph{Opportunity-load intervention.}
\label{sec:method_load}
The value of temporal allocation depends on both demand and capacity. At the
B10 PM-Bench operating point, we hold the 21-query cap fixed and vary the
number of latent-state trigger opportunities. The $k=0$ condition uses the
unmodified background; $k=3$ adds 21 opportunities while the horizon, query
prices, task construction, and paired background remain fixed. One Myopic and
one Sequential policy are trained across both loads and deployed unchanged
from observable pre-query state. We compare structural
headroom $H_k$ and learned gain
$G_k=Q_{\mathrm{Seq},k}-Q_{\mathrm{Myo},k}$ across the two loads, where $Q$
denotes mean held-out episode Set F1. Compiler,
weighting, and out-of-fold details appear in Appendix~\ref{app:compiler}.

\paragraph{Data splits.}
PM-Bench development, system/local analysis, confirmation, severe-scarcity,
and boundary studies use distinct seed ranges summarized in
Appendix~\ref{app:data_splits}. GoodAI groups complete trajectories for
out-of-fold fitting. The controlled-load study trains on 50 paired development
weeks and evaluates 240 paired held-out weeks.

\paragraph{Statistical analysis.}
We report paired 95\% bootstrap intervals~\citep{efron1993bootstrap} with
100,000 resamples unless noted otherwise. Each resampling unit is a complete
PM-Bench week/scenario or GoodAI trajectory. Paired load conditions are
resampled together by base seed. A prespecified power calculation fixes the
controlled-load sample size (Appendix~\ref{app:compiler}).

\section{Results}

\subsection{BudgetPM-Static leads complete-system comparisons across backbones}
\label{sec:system_results}
BudgetPM-Static achieves the highest end-to-end task quality in all six
backbone--benchmark combinations (Table~\ref{tab:system_baselines}). The gap
to the strongest Mem0/PMA workflow ranges from 0.362 to 0.452 Set F1 on
PM-Bench and from 0.433 to 0.920 accuracy on GoodAI. With Qwen3-14B, the
BudgetPM workflow reaches 0.885 PM-Bench Set F1 and 0.944 GoodAI accuracy.

\begin{table}[!h]
\centering
\caption{Complete-workflow task quality averaged over seeds 4000--4029.
BudgetPM uses Static at B60; Mem0 and PMA retain their adapted native workflows.
Controlled allocation tests follow; full protocol and GoodAI component scores
appear in Appendix~\ref{app:system_baselines}.}
\label{tab:system_baselines}
\small
\setlength{\tabcolsep}{4pt}
\begin{tabularx}{0.88\linewidth}{@{}ll>{\centering\arraybackslash}X>{\centering\arraybackslash}X@{}}
\toprule
Backbone & System & PM-Bench Set F1 & GoodAI Macro Acc. \\
\midrule
\multirow{3}{*}{Llama-3-8B}
& Mem0 & 0.201 & 0.000 \\
& PMA  & 0.398 & 0.026 \\
\rowcolor{staticrow}
& \textbf{\textsc{BudgetPM} (Static)} & \textbf{0.811} & \textbf{0.946} \\
\midrule
\multirow{3}{*}{Qwen2.5-14B}
& Mem0 & 0.484 & 0.450 \\
& PMA  & 0.490 & 0.513 \\
\rowcolor{staticrow}
& \textbf{\textsc{BudgetPM} (Static)} & \textbf{0.852} & \textbf{0.946} \\
\midrule
\multirow{3}{*}{Qwen3-14B}
& Mem0 & 0.433 & 0.500 \\
& PMA  & 0.432 & 0.500 \\
\rowcolor{staticrow}
& \textbf{\textsc{BudgetPM} (Static)} & \textbf{0.885} & \textbf{0.944} \\
\bottomrule
\end{tabularx}
\end{table}
The gate trained with Llama transfers to both Qwen backbones without retraining
or recalibration (Appendix~\ref{app:transfer}).

\subsection{A lightweight Outcome gate reaches the learned quality--cost frontier}
\label{sec:static_results}
Table~\ref{tab:confirmatory} asks whether Outcome gating requires a
high-capacity scorer. Static uses a single Logistic model. LambdaMART and
Ensemble use the same Outcome-supervised workflow with more expressive
scorers. At B40, Static achieves 0.814 F1 with 3.53 fewer queries than
LambdaMART and 5.56 fewer than Ensemble. At B60/B80, its F1 remains within
0.002 of LambdaMART. Static also retains 99.9\% of unconstrained
\emph{All relevant} F1 at B60 with 54.1\% fewer queries and preserves the full
0.820 F1 at B80 with 41.9\% fewer queries.

\begin{table}[h]
\centering
\caption{Matched PM-Bench local-selection controls on 30 paired weeks
(seeds 6000--6029). Row names identify the selection signal and scorer.
Each budget reports mean Set F1 ($\uparrow$) and realized query cost
($\downarrow$); unconstrained \emph{All relevant} achieves 0.820 F1 at cost
195.03.}
\label{tab:confirmatory}
\small
\setlength{\tabcolsep}{3pt}
\begin{tabularx}{0.98\linewidth}{>{\raggedright\arraybackslash}p{0.27\linewidth}*{8}{>{\centering\arraybackslash}X}}
\toprule
& \multicolumn{2}{c}{B20}
& \multicolumn{2}{c}{B40}
& \multicolumn{2}{c}{B60}
& \multicolumn{2}{c}{B80} \\
\cmidrule(lr){2-3}\cmidrule(lr){4-5}\cmidrule(lr){6-7}\cmidrule(lr){8-9}
Selection / scorer
& F1 & Cost & F1 & Cost & F1 & Cost & F1 & Cost \\
\midrule
Relevance heuristic
& 0.592 & 12.33 & 0.592 & 12.47 & 0.785 & 114.30 & 0.806 & 138.37 \\
Matched Due + Logistic
& 0.773 & 34.37 & 0.808 & 60.97 & 0.816 & 89.27 & 0.819 & 115.27 \\
Outcome + LambdaMART
& 0.782 & 35.23 & 0.812 & 64.90 & 0.821 & 88.83 & 0.820 & 109.73 \\
Outcome + Ensemble
& 0.761 & 31.67 & 0.809 & 66.93 & 0.819 & 88.73 & 0.820 & 115.07 \\
\midrule
\rowcolor{staticrow}
\textbf{Outcome + Logistic (Static)}
& 0.771 & 31.17
& 0.814 & 61.37
& 0.819 & 89.47
& 0.820 & 113.23 \\
\bottomrule
\end{tabularx}
\end{table}
\FloatBarrier

The matched-cost analysis reaches the same conclusion
(Appendix~\ref{app:matched_cost}). LambdaMART leads Static by 0.0055 F1 at a
mean cost of 40 queries, while Static leads by 0.0037 at 60. Their gaps narrow
to 0.0021 and 0.0005 at 90 and 100 queries. Static and LambdaMART therefore
lead at different operating points. The lightweight Logistic gate captures the
main Outcome-gating trade-off; additional model capacity changes selected
points rather than producing a consistent gain. Matched Due supervision and
the capacity controls separately identify the roles of the supervision target
and scorer (Appendices~\ref{app:due_control}
and~\ref{app:reviewer_diagnostics}).

Static extends the quality--cost advantage to semantic verification. On GoodAI,
it reaches 0.946/0.951 accuracy at B60/B80, gains of 1.33/1.67 percentage
points over the task-specific Due gate while reducing mean query use by
0.24/0.44
(Table~\ref{tab:goodai_task_gate}). Full budget and matched-supervision
results appear in Appendices~\ref{app:supporting_diagnostics}
and~\ref{app:due_control}.

These results show that local-value gating preserves task quality across the
standard-budget range. We next test tighter budgets, where useful checks begin
to compete across time.

\subsection{Sequential improves observation efficiency over natural monitoring}
\label{sec:natural_schedule_results}
Under the same hard caps, Sequential exceeds Deadline-Outcome, the strongest
of the three natural schedules, by 1.92--2.58 Set F1 points at B5/B10 on both
tests. Its on-time recall gain is 2.28--2.94 points. Its mean F1 is also higher
than Pacing-LB in all four comparisons.

Among the tested operating points, Deadline-Outcome requires
1.50$\times$ as many observations as Sequential at B5 and
1.19--1.20$\times$ at B10 to match its mean Set F1 and on-time recall.
Periodic requires 3.49--4.70$\times$ as many observations
(Table~\ref{tab:natural_schedule_efficiency_main}). Sequential therefore
provides the same observed service with 33\% fewer observations at B5 and
16--17\% fewer at B10.

\begin{table}[h]
\centering
\caption{Observation cost required by each natural schedule to match
Sequential's Set F1 and on-time recall, relative to Sequential
($1.00\times$; lower is better).}
\label{tab:natural_schedule_efficiency_main}
\small
\setlength{\tabcolsep}{3.5pt}
\begin{tabular*}{\linewidth}{@{}ll@{\extracolsep{\fill}}ccc@{}}
\toprule
Test & Budget & Deadline & Deadline+Backoff & Periodic \\
\midrule
\multirow{2}{*}{Severe}
& B5  & 1.50$\times$ & 1.50$\times$ & 4.70$\times$ \\
& B10 & 1.20$\times$ & 1.40$\times$ & 4.08$\times$ \\
\midrule
\multirow{2}{*}{Boundary}
& B5  & 1.50$\times$ & 1.50$\times$ & 4.19$\times$ \\
& B10 & 1.19$\times$ & 1.39$\times$ & 3.49$\times$ \\
\bottomrule
\end{tabular*}
\end{table}

\subsection{Future-aware supervision explains the low-budget gain}
\label{sec:sequential_results}
\begin{figure}[H]
  \centering
  \includegraphics[width=0.97\textwidth]{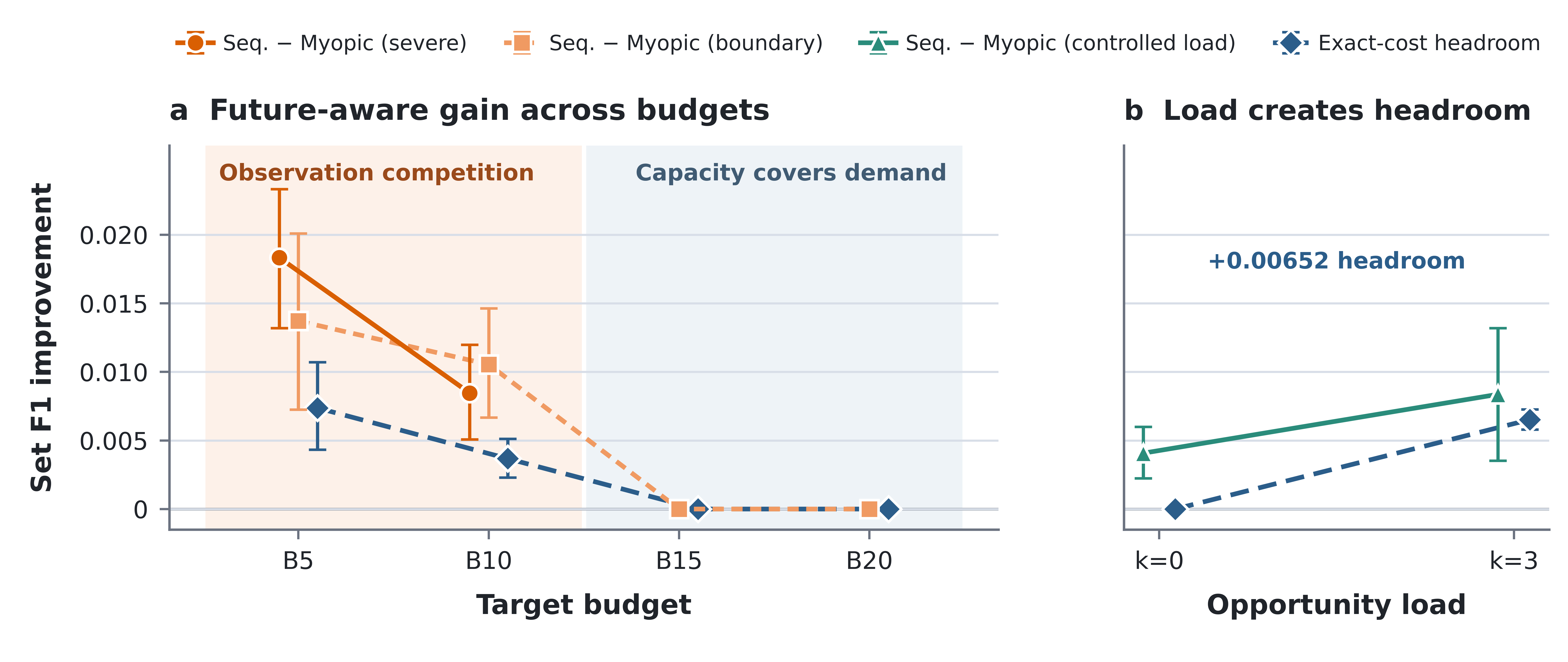}
  \caption{Future-aware gains follow the demand--capacity boundary. \textbf{a:}
  Sequential-minus-Myopic Set F1 gains on two disjoint tests and exact-cost
  structural headroom are positive at B5/B10, where observations compete, and
  zero at B15/B20, where capacity covers demand. \textbf{b:} At fixed B10,
  the controlled-load test shows that increasing opportunity load creates
  0.00652 exact-cost headroom while Sequential remains above Myopic. Error bars
  show paired-bootstrap 95\% CIs.}
  \label{fig:scarcity_boundary_load}
\end{figure}

Sequential improves over Myopic on both disjoint tests. The two policies differ
only in supervision, so this advantage measures the value of learning from
future-aware schedules
(Figure~\ref{fig:scarcity_boundary_load}a).

On boundary weeks 9000--9029, Static exhausts both the B5 and B10 budgets in
every week. After the budget is exhausted, 544 and 277 positive opportunities
remain, respectively. These opportunities are spread nearly uniformly across
episode quarters. This pattern explains why reserving capacity can create
exact-cost headroom and learned gains.

\subsection{Opportunity competition determines the scarcity boundary}
\label{sec:boundary_results}
At the same cost, hindsight DP exceeds current-greedy by $+0.00737$ F1 at B5
and $+0.00367$ at B10. Requiring the same number of queries at every step removes
the gap, attributing it to allocation across time
(Appendix~\ref{app:headroom_decomposition}). At B15/B20, the two teachers agree
on development data, so the deployed policies are identical. Held-out
exact-cost analysis also finds zero headroom
(Figure~\ref{fig:scarcity_boundary_load}a).

At fixed B10, increasing opportunity load from $k=0$ to $k=3$ raises exact-cost
structural headroom from 0 to $+0.00652$ F1. Under the same cap, the pooled
Sequential policy gains $+0.00407$ and $+0.00836$ over Myopic at the two loads.
The learned policies have their own realized costs
(Figure~\ref{fig:scarcity_boundary_load}b).

Together, the budget sweep and load intervention show that demand relative to
capacity---not the nominal budget alone---sets the boundary. Structural
headroom and Sequential's learned gain follow the same transition across
budgets. At a fixed budget, higher opportunity load creates new headroom, and
the learned gain moves in the same direction. The learning target should
therefore match the regime: local value when capacity covers useful checks, and
temporal allocation when present and future opportunities compete.

\section{Discussion and Conclusion}
BudgetPM treats the external-state checks required by stored intentions as a
resource-allocation problem. Static estimates current observation value;
Sequential learns how to distribute checks across an episode from future-aware
offline schedules. Both deploy as lightweight pre-query policies under a
shared hard-budget executor.

Across two benchmarks and three backbones, Static leads adapted Mem0/PMA
workflows. On PM-Bench, its lightweight Logistic scorer provides competitive
quality--cost operating points alongside LambdaMART and preserves
99.9--100\% of unconstrained quality with 42--54\% fewer observations. Under
severe scarcity and the same hard caps, Sequential exceeds the strongest
natural schedule by 1.92--2.58 Set F1 points. That schedule needs
$1.19$--$1.50\times$ as many observations to match Sequential's Set F1 and
on-time recall. Two disjoint matched tests also confirm Sequential's advantage
over Myopic.

The budget sweep and load intervention reveal a demand--capacity boundary:
local-value learning suffices when capacity covers useful checks, while
future-aware allocation adds value when present and future checks compete.
More broadly, stored intentions create future demand for external evidence.
BudgetPM establishes this memory--observation interface and shows why agent
memory systems should jointly manage stored intentions and the observations
needed to act on them.

\bibliography{budgetpm_references}
\bibliographystyle{iclr2027_conference}

\appendix
\section{Implementation and Data Construction}
\label{app:implementation}
This appendix specifies the data and learning procedures.
Appendix~\ref{app:static_evidence} follows the system and gating studies;
Appendix~\ref{app:temporal_evidence} reports the Sequential and scarcity
studies; Appendix~\ref{app:diagnostics} presents robustness and design evidence.

\subsection{Data splits and evaluation roles}
\label{app:data_splits}
Table~\ref{tab:splits} collects the PM-Bench splits used in this paper.
The standard-budget studies use the released V9 generator and its native
variation in horizon and task count. Controlled-load pairs use the frozen
paired generator described in Appendix~\ref{app:compiler}.
OOF folds keep each complete week/scenario, or paired base seed, intact.

\begin{table}[htbp]
\centering
\caption{PM-Bench split roles. A pair contains one $k=0$ and one $k=3$ week.}
\label{tab:splits}
\small
\setlength{\tabcolsep}{4pt}
\begin{tabular*}{0.80\linewidth}{@{}l@{\extracolsep{\fill}}ll@{}}
\toprule
Role & Seed range & Size \\
\midrule
Static and standard-budget development & 2000--2019 & 20 weeks \\
System comparison and Static analysis & 4000--4029 & 30 weeks \\
Static confirmation & 6000--6029 & 30 weeks \\
Severe-scarcity test & 8000--8029 & 30 weeks \\
Boundary confirmation & 9000--9029 & 30 weeks \\
Controlled-load development & 20000--20049 & 50 pairs \\
Controlled-load final test & 21000--21239 & 240 pairs \\
\bottomrule
\end{tabular*}
\end{table}

The PM-Bench BudgetPM pipeline retains the benchmark's state providers, action
space, and Set F1 scorer. The LLM receives visible natural-language headers,
updates, step text, and public channel names. It emits structured
create/update/cancel operations, followed by deterministic verification and
menu matching. The native system-baseline adapters are specified in
Appendix~\ref{app:system_baselines}.
GoodAI uses 12 development trajectories and evaluates seeds 4000--4029.

\subsection{Features, training, and calibration}
PM-Bench uses 27 pre-query features. Calendar and timing features encode
sinusoidal time, weekday, urgency, and the nearest signed and absolute time
distance. Resource and candidate features include resource cost, clock identity,
candidate counts, sum/max retrieval relevance, and channel resolvability. The
remaining features summarize legacy and calibrated utility, priority and miss
cost, unknown conditions, and trigger-type fractions. GoodAI adds seven features:
semantic-resource identity, semantic-trigger fraction, token and fuzzy trigger
similarity, mean memory age, repeat fraction, and active-memory count. Wall-clock
calendar features are neutralized for semantic resources because runner time is
not task evidence.

The nine temporal features for Myopic and Sequential are computed before each
decision. At one-indexed step $t$ of horizon $T$, with cap $B$, remaining
budget $R_t$, and target fraction $b$, the first six features are $b$; $R_t/B$;
$(t-1)/\max(1,T-1)$; $(T-t+1)/T$; $R_t/(T-t+1)$;
and $(B-R_t)/B-(t-1)/\max(1,T-1)$. The final three are query counts over the
previous five and ten steps divided by 10 and 20, respectively, and the
previous step's query count divided by two. Empty spend histories contribute
zero. The last three normalizers reflect PM-Bench's two-query per-step cap.

\paragraph{Static fitting and calibration.}
Static uses standardized Logistic regression~\citep{pedregosa2011scikit} with
inverse regularization strength $C=1$, L2 regularization, \texttt{lbfgs},
balanced class weights, \texttt{max\_iter=5000}, and tolerance $10^{-4}$.
Experiments use Python 3.10.19, scikit-learn 1.7.2, and
NumPy 1.24.4~\citep{harris2020array}. Cross-fitting groups complete PM-Bench
weeks/scenarios into five folds (20 development weeks), and complete GoodAI
trajectories into four folds (12 development trajectories).
Equation~\ref{eq:calibration} selects Static thresholds from OOF scores to
match the target query count. The sigmoid output is a positive-label score,
not a calibrated posterior probability.

\begin{table}[htbp]
\centering
\caption{Frozen Static thresholds selected on development data.}
\label{tab:static_thresholds}
\small
\setlength{\tabcolsep}{4pt}
\begin{tabular*}{0.80\linewidth}{@{}l@{\extracolsep{\fill}}lrrrr@{}}
\toprule
Benchmark & Labels & B20 & B40 & B60 & B80 \\
\midrule
PM-Bench & Outcome & 0.51608 & 0.33971 & 0.12289 & 0.00283 \\
         & Due     & 0.62312 & 0.32715 & 0.15360 & 0.03700 \\
GoodAI   & Outcome & 0.78747 & 0.28710 & 0.08381 & 0.02749 \\
         & Due     & 0.80578 & 0.29739 & 0.08343 & 0.02729 \\
\bottomrule
\end{tabular*}
\end{table}

For the severe-budget Static references, B5 and B10 use the B20 threshold
clamp. Intermediate fractions use linear threshold interpolation.

\paragraph{Temporal-policy fitting under severe scarcity.}
Myopic and Sequential use the same standardized Logistic family with $C=1$,
\texttt{lbfgs}, balanced class weights, and \texttt{max\_iter=4000}.
Their thresholds are selected from $0.05,0.075,\ldots,0.95$ to maximize mean
complete-week OOF Set F1. Ties favor lower realized cost and then the higher
threshold. This quality-based selection differs from Static's target-count
calibration.

\paragraph{Pooled fitting across opportunity loads.}
Automatic class weighting is disabled. Within each teacher dataset, let
$n_{k,y}$ count rows with condition $k\in\{0,3\}$ and teacher membership label
$y\in\{0,1\}$. Each row receives raw weight $1/(4n_{k,y})$, rescaled so the
mean row weight is one. Thus the four condition--label cells have equal total
weight. Standardization uses the same row weights. Paired load variants remain
in the same OOF fold. One threshold per teacher is selected from the same
grid by equal-weight mean complete-week OOF F1 across the two loads.
The selected Myopic/Sequential thresholds are 0.775/0.725 and are shared
across $k=0$ and $k=3$.

\paragraph{Executor settings and pre-query features.}
Every PM-Bench method uses the same per-step cap of two queries. We selected
this value on the released reference-week development split and kept it fixed
for all multi-week comparisons. Unit-cost resources are ordered by score. The
general executor uses the heuristic score/cost order.

All gate features are constructed before any state provider is invoked.
Priority, miss cost, confidence, actions, and trigger structure are internal memory
fields emitted by the LLM compiler from visible instructions and updates; they are
not copied from scenario task records or the scorer. Candidate relevance uses these
stored fields and the current visible observation. Urgency and time distance are
deterministic functions of the stored trigger and current timestamp. Channel
resolvability, unknown-condition counts, and trigger-type fractions inspect only
the stored trigger graph and public channel schema. ``Legacy'' and ``calibrated''
utility are deterministic combinations of relevance, compiler-provided risk fields,
urgency, and resolvability. The engine calls a selected channel provider only after the feature
rows have been returned and the gate has committed its plan. Consequently, none of the
27 features contains the current query result, a future observation, the due set, or
the local-effect label.

The timestamp is visible step metadata used to form pre-query calendar and
urgency features. It helps the allocator decide when to query. Selecting the
clock resource adds its result to the controller evidence and consumes budget.

The upstream PM-Bench state-channel actionability gate is fixed at 0.005 across
all matched traces. It defines the compiler-side candidate interface and is
distinct from BudgetPM's budget-specific threshold. The semantic-verifier
pipeline analogously uses a fixed 0.05 gate.

\subsection{Outcome-label construction by frozen replay}
\label{app:outcome_labels}
PM-Bench contains 4,332 candidate-resource rows over 20 development weeks/scenarios;
GoodAI contains 670 rows over 12 development trajectories. PM-Bench's collection
scheduler snapshots every relevant provider without committing state and enumerates
every affordable subset. Its Outcome target uses the singleton-versus-empty
comparison in Equation~\ref{eq:local_effect}. GoodAI collects every candidate
semantic query and uses the leave-one-out deletion comparison in the same equation.
All provider/verifier acquisition occurs in the source rollout. Subsequent label
enumeration, teacher construction, and DP reuse cached outcomes and deterministic
recomposition, with no additional LLM calls. The resulting scorers and thresholds
are fitted once and reused across deployment episodes.

Scenario steps and external channel contents are fixed by the generated seed.
Observation choices update query history and remaining budget, while executed
actions update task- and memory-completion state. Outcome effects are computed
under deterministic frozen recomposition, with the stored compiler/base
response, provider/verifier results, candidate set, and benchmark scorer held
fixed. For structural analysis, DP operates on the fixed full-subset source
trajectory and opportunity sequence. Myopic and Sequential deployment runs
instead advance their own online budget and history states.

\section{System Comparisons and Local Observation Gating}
\label{app:static_evidence}

\subsection{Complete-system baseline protocol}
\label{app:system_baselines}
The system-level comparison in Table~\ref{tab:system_baselines} uses the same 30
held-out seeds (4000--4029) for every backbone--system pair. We average PM-Bench
Set F1 and GoodAI macro accuracy over seeds. The prospective-memory and
trigger-response components are also reported separately. Mem0 and Proactive
Memory Agent provide public end-to-end memory-agent baselines; matched BudgetPM
controls provide mechanism-level comparisons.

\paragraph{Comparison scope and execution settings.}
The Mem0/PMA adapters supply memory context to the benchmark agent, which selects
actions through the official PM-Bench interface. These runs retain the benchmark
tool-request limits and have no BudgetPM episode-level observation cap or learned
observation gate. BudgetPM uses its Static policy with the B60 threshold and hard-cap executor. Thus the
system comparison evaluates complete workflows, while the shared-controller
comparisons in Sections~\ref{sec:static_results} and~\ref{sec:sequential_results}
test gating and future-aware supervision, respectively.

Mem0 uses its memory extraction and retrieval pipeline; PMA uses its proactive
memory pipeline. The Llama-3-8B adapters use a fixed context-compaction rule to fit
the 8192-token window and a compact, schema-compatible Mem0 extraction prompt.
The frozen rule preserves system instructions, task anchors, and recent turns,
then shortens injected memory context when necessary. PM-Bench action completions
are capped at 768 tokens. These implementation settings are part of the evaluated
system adapters.

\begin{table}[h]
\centering
\caption{Complete Mem0 and PMA quality matrix on seeds 4000--4029. The GoodAI column reports macro accuracy with prospective-memory / trigger-response component accuracy in parentheses.}
\small
\setlength{\tabcolsep}{4pt}
\begin{tabular*}{0.82\linewidth}{@{}l@{\extracolsep{\fill}}lcc@{}}
\toprule
Backbone & System & PM-Bench Set F1 & \shortstack{GoodAI Macro Acc.\\(Pros. / Trigger)} \\
\midrule
\multirow{2}{*}{Llama-3-8B}
& Mem0 & 0.201 & 0.000 (0.000/0.000) \\
& PMA & 0.398 & 0.026 (0.013/0.038) \\
\midrule
\multirow{2}{*}{Qwen2.5-14B}
& Mem0 & 0.484 & 0.450 (0.013/0.887) \\
& PMA & 0.490 & 0.513 (0.027/1.000) \\
\midrule
\multirow{2}{*}{Qwen3-14B}
& Mem0 & 0.433 & 0.500 (0.000/1.000) \\
& PMA & 0.432 & 0.500 (0.013/0.987) \\
\bottomrule
\end{tabular*}
\end{table}

\subsection{Backbone transfer without retraining}
\label{app:transfer}
The transfer evaluation covers Llama-3-8B~\citep{grattafiori2024llama},
Qwen2.5-14B~\citep{qwen2025qwen25}, and Qwen3-14B~\citep{yang2025qwen3}. The gate
learned with the primary Llama-3-8B pipeline transfers without retraining or
threshold recalibration; the agent backbone is the sole changed component.

\begin{table}[htbp]
\centering
\caption{Zero-adaptation transfer. Each cell is quality / mean query cost over 30
held-out seeds 4000--4029. Qwen3 reports B60, the complete-system operating point.}
\label{tab:scale}
\small
\setlength{\tabcolsep}{4pt}
\begin{tabular*}{0.86\linewidth}{@{}l@{\extracolsep{\fill}}rrrr@{}}
\toprule
& \multicolumn{2}{c}{PM-Bench} & \multicolumn{2}{c}{GoodAI} \\
Model & B40 & B60 & B40 & B60 \\
\midrule
Llama-3-8B & 0.802 / 62.97 & 0.811 / 90.90 & 0.927 / 20.17 & 0.946 / 30.13 \\
Qwen2.5-14B & 0.821 / 76.63 & 0.852 / 105.10 & 0.927 / 20.17 & 0.946 / 30.13 \\
Qwen3-14B & -- & 0.885 / 121.87 & -- & 0.944 / 30.13 \\
\bottomrule
\end{tabular*}
\end{table}

For Qwen3-14B at B60, the mean PM-Bench Set F1 is 0.8847
(95\% CI [0.8717, 0.8957]); GoodAI accuracy is 0.9444
[0.9244, 0.9622]. GoodAI prospective-memory and trigger-response accuracies
are 0.9867 and 0.9022. The same frozen Static gate and thresholds are
used for every backbone, demonstrating zero-adaptation transfer across model
families.

\subsection{Static gating at matched realized cost}
\label{app:matched_cost}

To compare static methods at similar realized query cost, we evaluate each fixed
policy over 13 development-derived operating points and interpolate F1 at
matched mean costs. The comparison separates the selection signal from scorer
capacity. \emph{Relevance} uses the retrieval score directly. The
\emph{Task-specific Due gate} follows its original Logistic training and
calibration recipe. \due and Static are matched Logistic gates trained with Due
and Outcome labels, respectively. The matched LambdaMART
\citep{burges2010ranknet} and hurdle/pairwise ensemble retain Outcome
supervision and increase scorer capacity.

\begin{table}[h]
\centering
\caption{Interpolated F1 at fixed mean realized-query costs on held-out PM-Bench
weeks 6000--6029. The 13 operating points are fixed from development data;
values interpolate adjacent points on each frozen policy curve.}
\label{tab:matched_cost}
\small
\setlength{\tabcolsep}{4pt}
\begin{tabularx}{0.78\linewidth}{@{}l*{4}{>{\centering\arraybackslash}X}@{}}
\toprule
Method & 40 & 60 & 90 & 100 \\
\midrule
Relevance heuristic & 0.6643 & 0.7168 & 0.7722 & 0.7774 \\
Task-specific Due/Logistic & 0.7690 & 0.7877 & 0.8070 & 0.8103 \\
Matched Due/Logistic & 0.7818 & 0.8067 & 0.8165 & 0.8191 \\
Outcome/LambdaMART & 0.7897 & 0.8090 & 0.8211 & 0.8205 \\
Outcome/Ensemble & 0.7761 & 0.8018 & 0.8194 & 0.8201 \\
\midrule
\rowcolor{staticrow}
\textbf{Outcome/Logistic (Static)} & 0.7842 & 0.8127 & 0.8190 & 0.8200 \\
\bottomrule
\end{tabularx}
\end{table}

At matched realized costs, Static improves over the task-specific Due gate at
all four reported costs. At 60 queries, the gain over this gate is 0.0250 F1.
Across the full 13-point sweep, Static also contributes quality--cost frontier
points at target settings B20--B55. Static and LambdaMART lead at different costs:
LambdaMART is higher by 0.0055 F1 at 40 queries, whereas Static is higher by
0.0037 at 60. Their differences narrow to 0.0021 and 0.0005 at 90 and 100
queries. Thus, the lightweight Logistic gate reaches competitive operating
points without relying on the higher-capacity scorer.

The corresponding GoodAI comparison uses seeds 4000--4029 and the semantic
query interface. Table~\ref{tab:goodai_task_gate} shows higher accuracy for
Static at B60/B80 with slightly lower realized cost. The task-specific Due gate
uses its original training and calibration recipe, so this table compares
complete gates rather than isolating the supervision label.

\begin{table}[h]
\centering
\caption{GoodAI gate comparison on 30 paired analysis trajectories. The first
two rows give macro accuracy / mean realized semantic-verifier queries; the
last gives the accuracy gain computed before rounding.}
\label{tab:goodai_task_gate}
\small
\setlength{\tabcolsep}{4pt}
\begin{tabular*}{0.68\linewidth}{@{}l@{\extracolsep{\fill}}rr@{}}
\toprule
Method & B60 & B80 \\
\midrule
Task-specific Due gate & 0.932 / 30.37 & 0.934 / 40.07 \\
Static & \textbf{0.946} / 30.13 & \textbf{0.951} / 39.63 \\
\midrule
Static gain (pp) & +1.33 & +1.67 \\
\bottomrule
\end{tabular*}
\end{table}

\subsection{Random-feasible and unconstrained baselines}
\label{app:random_references}

\emph{Random feasible} samples from the candidate queries that are currently
feasible under the same per-step and episode caps. \emph{All relevant} on
PM-Bench and \emph{All semantic} on GoodAI execute every candidate query after
candidate generation. They use neither the episode cap nor the per-step cap and
serve as unconstrained references.

\begin{table}[htbp]
\centering
\caption{Random feasible selection and unconstrained query references. Random
entries average three repetitions per seed (PM-Bench: confirmation seeds
6000--6029; GoodAI: seeds 4000--4029) and are
quality / mean realized cost. The unconstrained row has no target budget.}
\small
\setlength{\tabcolsep}{4pt}
\begin{tabular*}{0.90\linewidth}{@{}l@{\extracolsep{\fill}}lrrrr@{}}
\toprule
Benchmark & Method & B20 & B40 & B60 & B80 \\
\midrule
\multirow{2}{*}{PM-Bench}
& Random feasible & 0.613 / 38.58 & 0.655 / 74.70 & 0.696 / 105.30 & 0.735 / 128.16 \\
& All relevant & \multicolumn{4}{c}{0.820 / 195.03 (unconstrained)} \\
\midrule
\multirow{2}{*}{GoodAI}
& Random feasible & 0.579 / 9.37 & 0.664 / 20.04 & 0.756 / 30.40 & 0.836 / 40.03 \\
& All semantic & \multicolumn{4}{c}{0.993 / 57.67 (unconstrained)} \\
\bottomrule
\end{tabular*}
\end{table}

Random selection trails learned gating throughout the budget range. On PM-Bench,
Static at B80 matches the unconstrained 0.820 Set F1 while using 41.9\% fewer
queries, quantifying the efficiency gained by selective observation.

\subsection{Matched Outcome--Due supervision control}
\label{app:due_control}

We report the matched label comparison defined in
Section~\ref{sec:supervision_control}. The gate architecture, development data,
calibration procedure, and executor are held fixed to isolate the effect of
Outcome versus Due-state supervision.

\begin{table}[htbp]
\centering
\caption{Matched Outcome-versus-Due control. Every cell uses identical data volume,
features, logistic capacity, calibration, executor, and hard cap. Entries are
quality / mean realized cost on analysis seeds 4000--4029.}
\small
\setlength{\tabcolsep}{4pt}
\begin{tabular*}{0.90\linewidth}{@{}l@{\extracolsep{\fill}}lrrrr@{}}
\toprule
Benchmark & Supervision & B20 & B40 & B60 & B80 \\
\midrule
\multirow{2}{*}{PM-Bench}
& Due state & 0.762 / 36.33 & 0.798 / 63.53 & 0.808 / 92.90 & 0.810 / 114.57 \\
& Outcome & 0.760 / 30.57 & 0.802 / 62.97 & 0.811 / 90.90 & 0.811 / 115.60 \\
\midrule
\multirow{2}{*}{GoodAI}
& Due state & 0.787 / 9.83 & 0.924 / 20.17 & 0.941 / 30.20 & 0.948 / 39.73 \\
& Outcome & 0.789 / 9.87 & 0.927 / 20.17 & 0.946 / 30.13 & 0.951 / 39.63 \\
\bottomrule
\end{tabular*}
\end{table}

At PM-Bench B20, Outcome supervision achieves 0.760 F1 with 5.76 fewer queries
than Due-state supervision. It improves F1 at B40--B80 and across GoodAI
budgets. This comparison identifies task effect as a useful gating target
beyond trigger need.

\FloatBarrier
\subsection{Remaining-budget pacing}
\label{app:supporting_diagnostics}

\emph{Pacing} is a variant of Static that keeps its Outcome scorer and hard-cap
executor. At each step, it scales the calibrated target fraction by the ratio
of remaining budget to remaining estimated horizon, clips the result to
$[0,1]$, and looks up the corresponding score threshold. PM-Bench uses the
exact scenario length. GoodAI uses the development-median horizon of 66
decisions. For fractions strictly between zero and
one, Pacing clamps threshold lookup to the nearest
20\% or 80\% endpoint outside its calibrated range and interpolates within it.
Fractions zero and one mean query none/all before hard-cap filtering. The
additional B5/B10-calibrated variant is specified in
Appendix~\ref{app:pacing_lowbudget}.

\begin{table}[htbp]
\centering
\caption{Static versus remaining-budget-paced Outcome gating. Each entry is
quality / mean realized cost on analysis seeds 4000--4029.}
\label{tab:pacing}
\small
\setlength{\tabcolsep}{4pt}
\begin{tabular*}{0.72\linewidth}{@{}l@{\extracolsep{\fill}}lrr@{}}
\toprule
Benchmark & Budget & Static & Pacing \\
\midrule
\multirow{4}{*}{PM-Bench}
 & B20 & 0.760 / 30.57 & 0.770 / 40.87 \\
 & B40 & 0.802 / 62.97 & 0.804 / 79.47 \\
 & B60 & 0.811 / 90.90 & 0.811 / 112.00 \\
 & B80 & 0.811 / 115.60 & 0.811 / 130.40 \\
\midrule
\multirow{4}{*}{GoodAI}
 & B20 & 0.789 / 9.87 & 0.790 / 9.93 \\
 & B40 & 0.927 / 20.17 & 0.928 / 20.63 \\
 & B60 & 0.946 / 30.13 & 0.967 / 30.40 \\
 & B80 & 0.951 / 39.63 & 0.971 / 40.07 \\
\bottomrule
\end{tabular*}
\end{table}

On PM-Bench, Static reaches the same B60/B80 quality as Pacing with 21.10/14.80
fewer queries. On GoodAI, Pacing raises B60/B80 accuracy to 0.967/0.971 with
less than half an additional query. Remaining-budget adaptation therefore
complements the shared executor when increased utilization benefits the task,
while Static supplies the lower-cost operating point.

\subsection{Interpolation to unseen budgets}
For this interpolation check, \emph{Relevance} is the PM-Bench retrieval
heuristic. \emph{Similarity} is the corresponding GoodAI trigger-text
heuristic. \emph{Task-specific Due gate} is the Logistic baseline
with its original training and calibration recipe.
\emph{Budget-tuned ens.} is the fixed-route, higher-capacity PM-Bench ensemble.
Static is the final Outcome-supervised local policy in BudgetPM.

\begin{table}[htbp]
\centering
\caption{Unseen-budget interpolation on analysis seeds 4000--4029. Each cell is mean quality / realized cost.}
\small
\setlength{\tabcolsep}{4pt}
\begin{tabular*}{0.84\linewidth}{@{}l@{\extracolsep{\fill}}lrrr@{}}
\toprule
Benchmark & Method & B30 & B50 & B70 \\
\midrule
\multirow{4}{*}{PM-Bench}
& Relevance & 0.580 / 14.70 & 0.753 / 81.43 & 0.793 / 133.03 \\
& Task-specific Due gate & 0.777 / 61.50 & 0.803 / 96.10 & 0.807 / 113.67 \\
& Budget-tuned ens. & 0.767 / 43.40 & 0.784 / 71.47 & 0.806 / 96.27 \\
& Static & \best{0.789} / 50.10 & \best{0.807} / 71.60 & \best{0.811} / 96.73 \\
\midrule
\multirow{3}{*}{GoodAI}
& Similarity & 0.637 / 12.23 & 0.677 / 24.97 & 0.672 / 35.73 \\
& Task-specific Due gate & \best{0.906} / 15.40 & \best{0.959} / 23.83 & 0.941 / 35.10 \\
& Static & 0.904 / 15.53 & 0.956 / 24.27 & \best{0.953} / 34.57 \\
\bottomrule
\end{tabular*}
\end{table}

Static leads at all three PM-Bench interpolation budgets. On GoodAI, it reaches
0.904/0.956/0.953 accuracy at B30/B50/B70 and leads at B70 by 0.012 while
using 0.53 fewer queries. The frozen calibration supports intermediate
operating points across both benchmarks.

\subsection{Outcome-label sample efficiency}
\label{app:label_efficiency}
This experiment quantifies how much replay supervision is needed to learn the
local gate. We fit on subsets of the development trajectories and evaluate on
the same 30 analysis seeds (4000--4029).
\begin{figure}[htbp]
  \centering
  \includegraphics[width=0.92\linewidth]{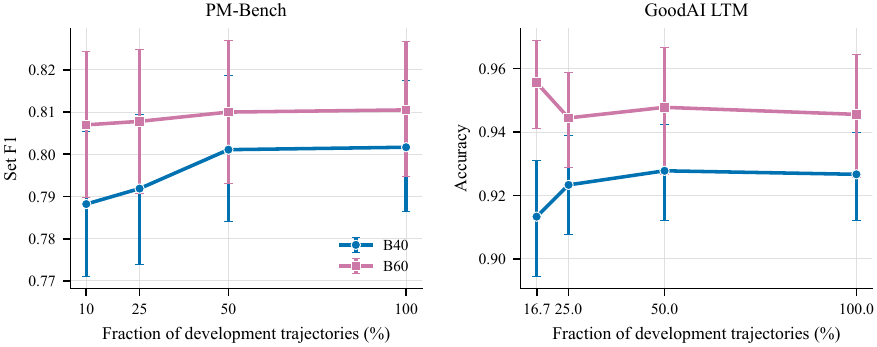}
  \caption{Held-out quality versus Outcome-label data. Error bars are paired
  bootstrap 95\% CIs over analysis seeds 4000--4029. GoodAI reaches an early performance plateau.}
  \label{fig:efficiency}
\end{figure}

\section{Future-Aware Allocation under Limited Capacity}
\label{app:temporal_evidence}

\subsection{Budget utilization and oracle-gap decomposition}
\label{app:sequential_opportunity}

Across the primary Llama-3-8B analysis and confirmation traces at B20--B80,
every above-threshold Static proposal fits within the episode cap. These budgets
form the local-selection regime identified in the main text: gating determines
efficiency, and the executor guarantees feasibility.

For the episode-level structural analysis, we replay complete per-step query
subsets and optimize aggregate episode Set F1 on the fixed full-subset recomposition
trajectory. DP is constrained to exactly match the reconstructed Static rule's
integer query cost. This analysis uses dedicated full-subset collection
trajectories; Section~\ref{sec:static_results} reports policy-induced Static
costs from deployment rollouts.

\begin{table}[h]
\centering
\caption{Exact-cost hindsight analysis on seeds 4000--4029, using
30 dedicated frozen full-subset collection trajectories. ``Reconstructed Static cost'' is evaluated on the collection trajectory;
``Later positives'' counts positive singleton opportunities strictly after the
reconstructed Static rule first exhausts the episode cap, averaged over weeks.
The total oracle gap can reflect both current-value ranking and temporal
allocation; it is not a direct measure of temporal headroom.}
\label{tab:exact_headroom}
\small
\setlength{\tabcolsep}{4pt}
\begin{tabular*}{0.88\linewidth}{@{}l@{\extracolsep{\fill}}rrrr@{}}
\toprule
Budget & \shortstack{Reconstructed\\Static cost} & \shortstack{Total oracle gap\\(DP--Static) [95\% CI]} & Exhaustion rate & Later positives \\
\midrule
B5  & 10.00  & +0.0457 [0.0406, 0.0508] & 1.00 & 17.60 \\
B10 & 21.00  & +0.0794 [0.0718, 0.0873] & 1.00 & 7.93 \\
B20 & 30.00  & +0.0741 [0.0680, 0.0804] & 0.00 & 0.00 \\
B40 & 56.93  & +0.0342 [0.0296, 0.0391] & 0.00 & 0.00 \\
B60 & 89.93  & +0.0253 [0.0208, 0.0299] & 0.00 & 0.00 \\
B80 & 115.50 & +0.0245 [0.0202, 0.0289] & 0.00 & 0.00 \\
\bottomrule
\end{tabular*}
\end{table}

A future-blind current-greedy oracle separates current-value ranking from
temporal allocation. At exact realized cost, DP-minus-current-greedy is
$+0.0063$ ($[0.0040, 0.0087]$) at B5 and $+0.0028$
($[0.0015, 0.0041]$) at B10. The difference is zero at B20/B40, where
current-greedy matches DP on the frozen recomposition problem. Positive singleton
opportunities are temporally diffuse across the analysis trajectories
(197/201/193/205 by quarter), as are the development Outcome counts
(106/108/94/83). These comparisons identify temporal allocation headroom at
B5/B10 and no headroom at B20/B40. The larger DP--reconstructed-Static gap also
reflects better ranking of current observations.

\subsection{Severe-scarcity policy comparison}
\label{app:learned_sequential}
The matched Myopic and Sequential policies are trained on 4,332 candidate rows from development weeks
2000--2019. The two teachers have matched global positive-label counts: 200 at B5
and 383 at B10. OOF selection yields Myopic/Sequential thresholds 0.950/0.250 at B5
and 0.850/0.875 at B10.

The final evaluation uses 30 held-out weeks disjoint from development and preceding
analysis weeks. All compared policies are evaluated on the same compiled task
trace within a week and share the same hard-budget executor. Table~\ref{tab:sequential_primary_intervals}
reports the paired severe-scarcity contrasts; Table~\ref{tab:severe_absolute}
gives absolute policy performance.

\begin{table}[h]
\centering
\caption{Primary sequential-minus-matched-myopic contrasts on weeks 8000--8029.
Intervals are paired complete-week bootstrap intervals with 100,000 resamples.}
\label{tab:sequential_primary_intervals}
\small
\setlength{\tabcolsep}{4pt}
\begin{tabular*}{0.82\linewidth}{@{}l@{\extracolsep{\fill}}rrrr@{}}
\toprule
Budget & $\Delta$F1 [95\% CI] & $\Delta$cost [95\% CI] & Wins & Ties \\
\midrule
B5  & +0.0183 [0.0132, 0.0233] & 0 [0, 0] & 25 & 2 \\
B10 & +0.0085 [0.0051, 0.0120] & $-0.23$ [$-0.47$, $-0.07$] & 19 & 9 \\
\bottomrule
\end{tabular*}
\end{table}

\begin{table}[htbp]
\centering
\caption{Absolute policy performance on severe-scarcity weeks 8000--8029. Entries are mean episode Set F1 / realized cost over 30 weeks. Pacing-LB is calibrated at B5/B10; Pacing retains the original standard-budget thresholds.}
\label{tab:severe_absolute}
\small
\setlength{\tabcolsep}{4pt}
\begin{tabular*}{0.62\linewidth}{@{}l@{\extracolsep{\fill}}rr@{}}
\toprule
Policy & B5 & B10 \\
\midrule
Static & 0.6225 / 10.00 & 0.6944 / 21.00 \\
Pacing & 0.6225 / 10.00 & 0.6944 / 21.00 \\
Pacing-LB & 0.6435 / 10.00 & 0.7159 / 20.93 \\
Periodic-Outcome & 0.5620 / 10.00 & 0.5874 / 20.73 \\
Myopic & 0.6367 / 10.00 & 0.7159 / 20.90 \\
Sequential & 0.6550 / 10.00 & 0.7243 / 20.67 \\
\bottomrule
\end{tabular*}
\end{table}

The corresponding spending profiles show that mean quarter-wise spend for
Static/Myopic/Sequential is 7.47/2.53/0.00/0.00, 5.67/3.70/0.13/0.50, and
3.70/5.30/1.00/0.00 at B5; at B10 it is 8.13/7.90/4.50/0.47,
6.17/6.80/5.97/1.97, and 5.43/6.43/5.67/3.13. Relative to Static, Sequential
imitation moves mean cap exhaustion from 0.311 to 0.519 episode progress at B5. At
B10, Static exhausts in every week at mean progress 0.653, whereas Sequential
imitation exhausts in 25/30 weeks and does so at mean progress 0.862 among those
weeks. At B5, Sequential reallocates first-quarter spend into the middle of the
episode; at B10, it extends allocation into the final quarter. These
budget-specific profiles show how Sequential redistributes queries across the
episode.

\subsection{Scarcity boundary across four budgets}
\label{app:boundary_confirmation}

The boundary confirmation evaluates B5/B10/B15/B20 on 30 disjoint weeks using the
same caps, outcome definitions, policy classes, and bootstrap procedure. Teacher agreement
on the development trajectories provides an important interpretation of the boundary:
current-greedy and hindsight-DP disagree on 310 steps (328 channel labels) at B5 and
16 steps at B10, and agree throughout B15/B20. The separately fitted Myopic and Sequential
policies are therefore identical at deployment at B15/B20, including their normalization,
coefficients, intercept, threshold, cap, and ordering.

Table~\ref{tab:boundary_primary_intervals} reports the paired learned contrasts,
reproducing the same B10--B15 scarcity boundary.

\begin{table}[h]
\centering
\caption{Primary boundary-confirmation contrasts on weeks 9000--9029. Intervals are
paired complete-week bootstrap intervals with 100,000 resamples.}
\label{tab:boundary_primary_intervals}
\small
\setlength{\tabcolsep}{4pt}
\begin{tabular*}{0.84\linewidth}{@{}l@{\extracolsep{\fill}}rrrr@{}}
\toprule
Budget & $\Delta$F1 [95\% CI] & $\Delta$cost [95\% CI] & Wins & Ties \\
\midrule
B5  & +0.0137 [0.0072, 0.0201] & 0 [0, 0] & 21 & 5 \\
B10 & +0.0105 [0.0067, 0.0146] & $-0.07$ [$-0.17$, 0] & 22 & 6 \\
B15 & 0 [0, 0] & 0 [0, 0] & 0 & 30 \\
B20 & 0 [0, 0] & 0 [0, 0] & 0 & 30 \\
\bottomrule
\end{tabular*}
\end{table}

\begin{table}[htbp]
\centering
\caption{Absolute policy performance on boundary weeks 9000--9029. Entries are mean episode Set F1 / realized cost over 30 weeks. Pacing-LB is calibrated at B5/B10; Pacing retains the original standard-budget thresholds.}
\label{tab:boundary_absolute}
\small
\setlength{\tabcolsep}{4pt}
\begin{tabular*}{0.82\linewidth}{@{}l@{\extracolsep{\fill}}rrrr@{}}
\toprule
Policy & B5 & B10 & B15 & B20 \\
\midrule
Static & 0.6236 / 10.00 & 0.6970 / 21.00 & 0.7459 / 29.57 & 0.7574 / 32.13 \\
Pacing & 0.6236 / 10.00 & 0.6970 / 21.00 & 0.7472 / 30.70 & 0.7669 / 41.23 \\
Pacing-LB & 0.6441 / 10.00 & 0.7151 / 21.00 & -- & -- \\
Periodic-Outcome & 0.5662 / 10.00 & 0.5975 / 20.63 & 0.6256 / 30.00 & 0.6461 / 36.37 \\
Myopic & 0.6340 / 10.00 & 0.7144 / 21.00 & 0.7486 / 29.23 & 0.7583 / 39.80 \\
Sequential & 0.6477 / 10.00 & 0.7250 / 20.93 & 0.7486 / 29.23 & 0.7583 / 39.80 \\
\bottomrule
\end{tabular*}
\end{table}

The temporal spending profiles show how Sequential redistributes queries across
the episode. At B5, Myopic spends 6.43/3.13/0.10/0.33 queries by quarter, whereas
Sequential spends 4.10/5.17/0.73/0. At B10 the corresponding profiles are
6.80/6.93/6.03/1.23 and 6.33/6.13/6.03/2.43; Sequential exhausts in 28/30 weeks at
mean progress 0.849 among exhausted weeks, versus 30/30 at 0.777 for Myopic. At
B15/B20 all timing summaries match exactly between the two learned policies, as
required by runtime identity.

\subsubsection{Natural monitoring schedules and quality--cost frontiers}
\label{app:periodic_control}
We compare Sequential with three deployment-feasible schedules. For a unit-cost
cap $K$, \emph{Periodic-Outcome} places $K$ opportunities uniformly over
normalized episode time and uses the frozen Static Outcome scorer to select the
highest-scoring unresolved resource. \emph{Deadline-Outcome} keeps this cadence
for state channels but prioritizes clock checks while an unresolved intention is
inside its declared time window. \emph{Deadline+Backoff-Outcome} uses the same
clock rule and exponentially increases the interval between unchanged state
checks, resetting after a change or newly relevant memory. All schedule inputs
are observable before querying, and all policies share the candidate space,
Outcome scorer, per-step limit, and hard-cap executor.

Backoff parameters are selected on analysis weeks 4000--4029 from initial
interval $\{1,2\}$, factor $\{1.5,2\}$, and maximum interval $\{4,8\}$. The
selection rule maximizes mean Set F1 across caps 10 and 21, with lower cost and
simpler parameters as tie breakers. It selects $(2,1.5,4)$. We then freeze all
controls and sweep 20 caps from 5 to 210 on the two disjoint test sets.

\begin{table}[htbp]
\centering
\caption{Sequential minus practical controls at matched B5/B10 caps.
Entries are Set F1 differences with paired 95\% CIs.}
\label{tab:scarcity_summary}
\small
\setlength{\tabcolsep}{4pt}
\begin{tabular}{llcc}
\toprule
Control & Test & B5 & B10 \\
\midrule
Pacing-LB & Severe & +0.0115 [0.0037, 0.0197] & +0.0084 [0.0047, 0.0121] \\
& Boundary & +0.0036 [$-0.0023$, 0.0096] & +0.0099 [0.0072, 0.0128] \\
Periodic & Severe & +0.0930 [0.0851, 0.1009] & +0.1370 [0.1260, 0.1481] \\
& Boundary & +0.0816 [0.0753, 0.0879] & +0.1275 [0.1196, 0.1353] \\
Deadline & Severe & +0.0258 [0.0204, 0.0314] & +0.0194 [0.0140, 0.0248] \\
& Boundary & +0.0192 [0.0123, 0.0259] & +0.0192 [0.0137, 0.0249] \\
Deadline+Backoff & Severe & +0.0303 [0.0245, 0.0361] & +0.0277 [0.0207, 0.0355] \\
& Boundary & +0.0218 [0.0139, 0.0295] & +0.0254 [0.0185, 0.0328] \\
\bottomrule
\end{tabular}
\end{table}

Deadline-Outcome is the strongest natural schedule at the matched operating
points. Table~\ref{tab:natural_schedule_service} asks a complementary question:
how much observation cost each control needs to reach both Sequential's mean
Set F1 and mean on-time recall. The deadline rule closes much of the gap to a
uniform cadence, yet Sequential reaches the same observed service point with
33\% fewer observations at B5 and 16--17\% fewer at B10.

\begin{table}[htbp]
\centering
\caption{Minimum observed cost that matches both Sequential's mean Set F1 and
on-time recall. Values summarize the 20-cap sweep.}
\label{tab:natural_schedule_service}
\small
\setlength{\tabcolsep}{4pt}
\begin{tabular}{llrrrr}
\toprule
Test & Target & Sequential & Deadline & Deadline+Backoff & Periodic \\
\midrule
Severe & B5  & 10.00 & 15.00 & 15.00 & 47.03 \\
& B10 & 20.67 & 24.80 & 29.00 & 84.40 \\
Boundary & B5  & 10.00 & 15.00 & 15.00 & 41.93 \\
& B10 & 20.93 & 24.93 & 29.10 & 73.13 \\
\bottomrule
\end{tabular}
\end{table}

\FloatBarrier
\subsubsection{Pacing calibrated for low budgets}
\label{app:pacing_lowbudget}
Pacing-LB retains the frozen Static Outcome scorer, the remaining-budget pacing
rule, and the common hard-cap executor. For each of B5 and B10, we add one
low-budget threshold knot to a separate copy of the frozen scorer; all other
knots remain unchanged. We search thresholds
$0.05,0.075,\ldots,0.95,0.975,0.99$ on the same 20 complete development weeks
2000--2019. Selection maximizes mean week-level Set F1, breaking ties by lower
mean realized cost and then a higher threshold. The selected knots are $0.99$
at B5 and $0.925$ at B10. Tuning changes only these two threshold knots; the
frozen Outcome scorer is selected on development weeks and applied unchanged to
both evaluation splits.

Each policy is evaluated on the same task trace and uses the same hard
cap within a test week.
Table~\ref{tab:scarcity_summary} reports the paired Sequential-minus-Pacing-LB
differences with 100,000 complete-week bootstrap resamples.
Tables~\ref{tab:severe_absolute} and \ref{tab:boundary_absolute} give absolute
quality and realized cost alongside Static and Myopic.

\subsubsection{Held-out exact-cost analysis}
On the same held-out weeks 9000--9029, we compute the exact-cost structural
comparison offline on the frozen full-subset recomposition trajectories without
refitting either learned policy. Hindsight DP is constrained in each week to exactly
the realized integer cost of the future-blind current-greedy oracle.

Figure~\ref{fig:scarcity_boundary_load}a summarizes these exact-cost comparisons
in the main text. The corresponding current-greedy / exact-cost-DP means are 0.6714/0.6788 at B5,
0.7832/0.7869 at B10, 0.8305/0.8305 at B15, and 0.8310/0.8310 at B20. The paired
DP-minus-current intervals are $+0.00737$ $[0.00433, 0.01071]$, $+0.00367$
$[0.00230, 0.00512]$, $0$ $[0, 0]$, and $0$ $[0, 0]$, respectively. Costs are matched
exactly at 10.00, 21.00, 26.50, and 26.57 queries. At B15/B20, current-greedy
matches the exact-cost DP quality; the few schedule substitutions preserve F1.

\paragraph{Decomposing temporal and within-step headroom.}
\label{app:headroom_decomposition}
To distinguish changes in query timing from changes in resource choice, we add
an exact restricted optimizer on the same 30 frozen boundary weeks. For each
week, let $S^{\mathrm{Fix}}$ maximize episode Set F1 while using exactly the
current-greedy number of queries at \emph{each step}. All resources have unit
cost here, so the restriction also fixes per-step cost. The exact-total-cost DP
can change that allocation across steps. Thus, on each week,
\[
Q_{\mathrm{DP}}-Q_{\mathrm{CG}}
=
\underbrace{Q_{\mathrm{DP}}-Q_{\mathrm{Fix}}}_{\text{across-step allocation}}
+
\underbrace{Q_{\mathrm{Fix}}-Q_{\mathrm{CG}}}_{\text{within-step choice}}.
\]
Table~\ref{tab:headroom_decomposition} reports the paired results. The
within-step component is exactly zero in every one of the 120 week--budget
cells. Positive B5/B10 headroom therefore comes entirely from permitting a
different query-count allocation across steps. This is a retrospective
decomposition of the frozen trajectories used for Figure~\ref{fig:scarcity_boundary_load}a;
the controlled-load trajectories in panel b are separate.

\begin{table}[h]
\centering
\caption{Exact-cost headroom decomposition on 30 frozen PM-Bench boundary weeks.
Brackets give paired-bootstrap 95\% confidence intervals for the across-step
component. Within-step choice is zero in every week.}
\label{tab:headroom_decomposition}
\small
\setlength{\tabcolsep}{4pt}
\begin{tabular*}{\linewidth}{@{}l@{\extracolsep{\fill}}ccc@{}}
\toprule
Budget & Full $\mathrm{DP}-\mathrm{CG}$ & Across-step allocation & Within-step choice \\
\midrule
B5  & +0.00737 & +0.00737 [0.00433, 0.01071] & 0 \\
B10 & +0.00367 & +0.00367 [0.00230, 0.00512] & 0 \\
B15 & 0 & 0 [0, 0] & 0 \\
B20 & 0 & 0 [0, 0] & 0 \\
\bottomrule
\end{tabular*}
\end{table}

The temporal analysis supports the same conclusion. Frozen positive singleton opportunities
are nearly uniform by episode quarter (193/197/204/207). Reconstructed Static
exhausts its cap in 100\%/100\%/60\%/13.3\% of weeks at B5/B10/B15/B20. Its
conditional mean exhaustion progress is 0.314/0.656/0.855/0.797, and positive
singleton opportunities after exhaustion total 544/277/72/26. Exact-cost
DP--current-greedy disagreements occur on 551/305/8/4 steps in 30/29/5/4 weeks.
By quarter, these counts are 179/100/91/181, 127/23/13/142, 3/0/0/5, and
1/0/0/3. Late opportunities remain present as allocation headroom collapses, linking
the boundary to available capacity rather than opportunity disappearance.

\subsection{Controlled-load study design}
\label{app:compiler}

\paragraph{Controlled-load construction.}
The controlled-load study uses the same frozen Llama-3-8B compiler in both
conditions. Holding the compiler fixed ensures that the intervention changes
observation demand rather than task interpretation.

\paragraph{Policy development.}
The controlled-load policies are trained on 50 paired development weeks
(seeds 20000--20049). The 36-feature representation excludes $k$ and condition
identity, and the two load variants of each base seed remain in the same OOF fold.
The two teachers agree throughout the $k=0$ development weeks and disagree in
all 50 high-load weeks, confirming that the added opportunities create temporal
allocation choices.

\paragraph{Final paired test.}
For base seed $i$ and load $k\in\{0,3\}$, let
$Q_{\mathrm{Seq},i,k}$ and $Q_{\mathrm{Myo},i,k}$ denote the corresponding
per-seed episode qualities, and define the per-seed learned gain
$G_{i,k}=Q_{\mathrm{Seq},i,k}-Q_{\mathrm{Myo},i,k}$. Let
$\rho=\operatorname{Corr}(G_{i,0},G_{i,3})$ denote the Pearson correlation
across paired base seeds between the two load-specific learned gains. We evaluate
240 paired held-out seeds, which provides at least 80\% two-sided power for a
0.005 F1 interaction under the prespecified correlation $\rho=-0.25$. The
compiler, policies, thresholds, hard-cap executor, and paired bootstrap
procedure are fixed across both load conditions.
The resulting learned and structural contrasts are reported in
Table~\ref{tab:paired_clean}; absolute performance is in
Table~\ref{tab:controlled_absolute}. Sequential uses approximately 0.95 and 0.90
more queries than Myopic at $k=0$ and $k=3$, with F1 gains of 0.00407 and
0.00836, respectively. Every policy remains within the same hard cap.

\begin{table}[h]
\centering
\caption{Paired controlled-load evaluation at B10 on 240 held-out base seeds. Learned entries are Sequential minus Myopic. Structural headroom is exact-cost DP minus the future-blind current-greedy oracle. Brackets give paired bootstrap 95\% confidence intervals.}
\label{tab:paired_clean}
\small
\setlength{\tabcolsep}{4pt}
\begin{tabular*}{0.82\linewidth}{@{}l@{\extracolsep{\fill}}cc@{}}
\toprule
Condition & \shortstack{Learned $\Delta$F1\\(policy costs)} & \shortstack{Structural $\Delta$F1\\(exact cost)} \\
\midrule
$k=0$ & +0.00407 [0.00224, 0.00599] & 0 [0, 0] \\
$k=3$ & +0.00836 [0.00353, 0.01320] & +0.00652 [0.00580, 0.00726] \\
\midrule
$I_L=G_3-G_0$ & +0.00429 [$-0.00064$, +0.00922] & -- \\
\bottomrule
\end{tabular*}
\end{table}

\begin{table}[htbp]
\centering
\caption{Absolute performance at fixed B10 on controlled-load test seeds
21000--21239. Entries are mean episode Set F1 / realized cost. Learned
policies are evaluated with their deployed histories; the two structural
references share frozen full-subset trajectories and are matched at exact cost.}
\label{tab:controlled_absolute}
\small
\setlength{\tabcolsep}{4pt}
\begin{tabular*}{0.68\linewidth}{@{}l@{\extracolsep{\fill}}rr@{}}
\toprule
Policy or structural reference & $k=0$ & $k=3$ \\
\midrule
Myopic & 0.7131 / 18.95 & 0.5670 / 19.96 \\
Sequential & 0.7171 / 19.90 & 0.5754 / 20.86 \\
\midrule
Current-greedy & 0.7384 / 14.58 & 0.6763 / 21.00 \\
Hindsight DP (exact cost) & 0.7384 / 14.58 & 0.6828 / 21.00 \\
\bottomrule
\end{tabular*}
\end{table}

At $k=0$, structural headroom is zero. The learned contrast follows each
policy's realized cost, whereas the structural comparison holds cost fixed on
the frozen trajectory.

\section{Robustness and Design Analyses}
\label{app:diagnostics}

\subsection{Supervision targets and heterogeneous costs}
\label{app:reviewer_diagnostics}

Table~\ref{tab:target_cost_diagnostics} compares supervision targets and query-price
profiles on analysis seeds 4000--4029, with B40/B60 hard caps and at most two
queries per step. Models and thresholds are selected using only the 20 development
trajectories. Binary Outcome/Logistic uses binary Outcome labels; alternatives use
magnitude weighting, hurdle expected gain, direct regression on $\Delta_{t,r}$,
or pairwise gain/cost. Heterogeneous-price rows divide each score by query cost.
Mild costs assign $(0.5,1,2)$ to clock, snapshot, and delta resources; skewed costs
assign $(0.25,1,4)$. Episode caps scale with each profile's maximum feasible
development cost.

\begin{table}[htbp]
\centering
\caption{Marginal-target and heterogeneous-cost comparisons. Each cell is PM-Bench
F1 / mean realized cost on analysis seeds 4000--4029.}
\label{tab:target_cost_diagnostics}
\small
\setlength{\tabcolsep}{4pt}
\begin{tabular*}{0.82\linewidth}{@{}l@{\extracolsep{\fill}}lrr@{}}
\toprule
Cost profile & Target / scorer & B40 & B60 \\
\midrule
\multirow{5}{*}{Unit}
& Binary Outcome / Logistic & 0.802 / 62.83 & 0.811 / 90.90 \\
& Magnitude-weighted Outcome / Logistic & 0.802 / 62.77 & 0.811 / 91.23 \\
& Hurdle expected gain & 0.800 / 62.97 & 0.809 / 93.27 \\
& Direct effect / Ridge & 0.776 / 66.70 & 0.797 / 94.63 \\
& Pairwise gain / cost & 0.804 / 68.83 & 0.809 / 95.77 \\
\midrule
\multirow{3}{*}{Mild}
& Relevance / cost & 0.765 / 83.48 & 0.794 / 125.58 \\
& Binary Outcome / Logistic / cost & 0.801 / 85.83 & 0.813 / 126.05 \\
& Hurdle expected gain / cost & 0.800 / 85.45 & 0.813 / 127.88 \\
\midrule
\multirow{3}{*}{Skewed}
& Relevance / cost & 0.794 / 152.01 & 0.793 / 181.78 \\
& Binary Outcome / Logistic / cost & 0.801 / 145.22 & 0.809 / 205.47 \\
& Hurdle expected gain / cost & 0.804 / 147.18 & 0.809 / 206.79 \\
\bottomrule
\end{tabular*}
\FloatBarrier
\end{table}

The Binary Outcome/Logistic row here and the Static row in
Table~\ref{tab:pacing} use separately frozen development-selected thresholds,
yielding B40 costs of 62.83 and 62.97, respectively.

Under unit costs, binary Outcome supervision matches the quality of richer
targets with fewer queries: pairwise gain/cost uses 6.00 and 4.87 additional
queries at B40 and B60. Under heterogeneous costs, Binary Outcome/Logistic
improves over Relevance/cost by $+0.0358$ and $+0.0188$ F1 under mild costs,
and by $+0.0069$ and $+0.0157$ under skewed costs.
These results support binary Outcome supervision as a simple quality--cost
choice across all three price profiles.

\subsection{Pairwise resource-interaction analysis}

For every PM-Bench development decision $t$ with at least two distinct candidate
resources $r,s\in\mathcal R_t$, we compute the local pairwise interaction
\begin{equation}
J_t(r,s)=m_t(\{r,s\})-m_t(\{r\})-m_t(\{s\})+m_t(\varnothing).
\label{eq:pairwise_interaction}
\end{equation}
Across 1,383 decisions, we evaluate 4,709 candidate-resource pairs.
Interactions are exactly zero in 99.47\% of cases; the remaining 0.53\% are
nonpositive. The median and 90th-percentile absolute interaction are zero, and
the maximum is 0.333. This structure supports the singleton-based Outcome
supervision used on PM-Bench.

\end{document}